\documentclass[11pt]{article}

\usepackage[final]{acl}

\usepackage{times}
\usepackage{latexsym}

\usepackage[T1]{fontenc}

\usepackage[utf8]{inputenc}

\usepackage{microtype}

\usepackage{inconsolata}

\usepackage{graphicx}

\usepackage{algorithm}
\usepackage{algpseudocode}
\usepackage{booktabs}
\usepackage{tcolorbox}
\usepackage{float}
\usepackage{amssymb}
\usepackage{colortbl}
\usepackage{multirow} 
\usepackage{footmisc}
\usepackage{tabularx}

\usepackage{adjustbox}
\usepackage{pifont}
\newcommand{\cmark}{\ding{51}} 
\newcommand{\xmark}{\ding{55}} 

\usepackage{amsmath,amssymb} 
\newcommand{\up}[1]{{\scriptsize(\(\uparrow\)#1)}}
\newcommand{\down}[1]{{\scriptsize(\(\downarrow\)#1)}}

\title{Learning When to Sample: Confidence-Aware Selective Sampling for Efficient Chain-of-Thought Reasoning}

\author{
 \textbf{Juming Xiong\textsuperscript{1}},
 \textbf{Kevin Guo\textsuperscript{1}},
 \textbf{Congning Ni\textsuperscript{2}},
 \textbf{Weixin Liu\textsuperscript{1}},
 \textbf{Chao Yan\textsuperscript{2}},
\\
 \textbf{Katherine Brown\textsuperscript{2}},
 \textbf{Avinash Baidya\textsuperscript{3}},
 \textbf{Xiang Gao\textsuperscript{3}},
 \textbf{Bradley Malin\textsuperscript{1,2}},
 \textbf{Zhijun Yin\textsuperscript{1,2}},
\\
\\
 \textsuperscript{1}Vanderbilt University,
 \textsuperscript{2}Vanderbilt University Medical Center,
 \textsuperscript{3}Intuit AI Research
\\
}

\begin{document}
\maketitle
\begin{abstract}

Large language models (LLMs) can achieve strong reasoning performance through chain-of-thought (CoT) reasoning, yet they often generate unnecessarily long reasoning paths that incur high inference cost. Self-consistency-based approaches push accuracy higher still, but they require sampling and aggregating multiple reasoning trajectories, leading to substantial computational overhead. In this paper, we introduce a confidence-aware selective sampling framework that, at inference time, analyzes a single reasoning trajectory to adaptively determine whether to rely on that trajectory alone or trigger multi-path sampling. The framework uses trajectory-level numeric and sentence-level linguistic features extracted from reasoning states to guide selective multi-path reasoning. We train it on MedQA and evaluate it in-domain on MedQA and under calibration-only transfer on MathQA, MedMCQA, and MMLU, without further fine-tuning. Experimental results show that the proposed framework maintains comparable performance to full and efficient multi-path reasoning baselines, with accuracy changes of $-0.41\pm0.58$ and $-0.31\pm0.58$ percentage points, respectively, while reducing token usage by $71.7\pm5.0\%$ and $36.6\pm9.1\%$. These findings demonstrate that reasoning trajectories contain rich signals for uncertainty estimation, enabling a simple, transferable mechanism to balance accuracy and efficiency in LLM reasoning.

\end{abstract}

\section{Introduction}
Large language models (LLMs) have recently demonstrated strong reasoning capabilities, enabling effective performance on a wide range of tasks, including mathematical problem-solving, commonsense reasoning, and scientific question answering \cite{wei2022cot,kojima2022zeroshot,sprague2025cotcotchainofthoughthelps}. This emerging reasoning ability represents an important step toward general-purpose problem-solving systems, as LLMs can decompose complex problems and produce coherent intermediate solutions \cite{wei2022emergent,yao2023tree}. Despite these promising observations, LLM reasoning remains unstable. A suboptimal local decoding decision can disrupt subsequent reasoning steps and propagate through the sequential generation trajectory \cite{peng2025stepwisereasoningerrordisruption,song2026large}. This phenomenon limits the reliability of single-pass decoding, suggesting that a single generation is often insufficient for robust problem solving  \cite{wang2023selfconsistency,song2026large}.

To address this gap, prior work has introduced various sampling-based aggregation methods. One representative method is self-consistency \cite{wang2023selfconsistency}, which samples multiple chain-of-thought (CoT) paths and then aggregates them through mechanisms such as majority voting and probabilistic consensus. It has been widely shown that self-consistency can substantially enhance robustness and accuracy compared to its single-path counterpart \cite{wang2023selfconsistency,xue2023dynamic}. However, these gains often require substantially more test-time computation, through longer outputs or repeated reasoning-path generation \cite{snell2025scaling,wu2025inference,muennighoff2025s1}. Such increased generation directly raises inference cost and energy consumption \cite{samsi2023wordstowatts,fernandez2025energy}, making the balance between reasoning performance and inference efficiency a fundamental challenge.

Motivated by this concern, recent work on efficient reasoning has explored adaptive test-time computation and selective sampling strategies that reduce redundant reasoning paths. Dynamic voting, for instance, reduces inference cost by terminating multi-sampled reasoning once sufficient agreement is reached across sampled chains \cite{xue2023dynamic}. More recent adaptive self-consistency methods further incorporate reasoning-path quality or response-level reliability signals to decide when additional samples are needed \cite{wan2025reasoning,kim2026reliability}. Broader test-time scaling studies also show that the amount of inference compute should be allocated adaptively rather than uniformly across all instances \cite{wu2025inference,zhang2025testtimescaling,alomrani2025reasoning}. Despite these efficiency gains, most methods still rely on observing multiple sampled trajectories or response-level agreement before deciding to stop, which incurs nontrivial computational overhead \cite{xue2023dynamic,wan2025reasoning,kim2026reliability}. As a result, the stopping decision may vary across runs and requires generating multiple responses, rather than relying on a single reasoning trajectory \cite{kadavath2022lmknow,yona2024faithfully}.

In this work, we propose a confidence-aware selective sampling framework that evaluates a completed single CoT reasoning trajectory to control inference-time reasoning cost. Specifically, our framework estimates the model's confidence and analyzes linguistic patterns of the initial reasoning trajectory, then adaptively determines whether to accept the answer deemed likely correct or trigger additional sampling for one assessed as likely wrong.
We evaluate the proposed framework on five popular LLMs: GPT-OSS 20B \cite{agarwal2025gpt}, Llama 3.1 8B Instruct \cite{dubey2024llama}, Qwen 2.5 7B \cite{qwen2025qwen25technicalreport}, and Qwen 3 14B/32B \cite{yang2025qwen3technicalreport}. The experiments are conducted across four multiple-choice benchmarks: MedQA \cite{jin2021medqa}, MathQA \cite{amini2019mathqa}, MedMCQA \cite{pal2022medmcqa}, and MMLU \cite{hendrycks2020measuring}. The results demonstrate that the proposed framework substantially improves inference efficiency, yielding major token savings, while preserving strong reasoning accuracy. In summary, our contributions are threefold:

\begin{itemize}

\item We introduce a confidence-aware selective sampling framework that analyzes a single completed CoT to determine whether additional multi-path reasoning is necessary, avoiding unnecessary sampling and computational overhead.

\item We define an attention-based recurrent neural network (RNN) decision model that leverages sentence-level numeric and linguistic features to capture temporal reasoning dynamics and assess the reasoning reliability.

\item We demonstrate the generalizability and robustness of the proposed approach through extensive evaluation across multiple LLMs and benchmarks, and provide ablation evidence that numeric and linguistic features provide complementary signals.

\end{itemize}

\section{Related Work}

\subsection{Reasoning in Large Language Models}
LLMs have demonstrated substantial improvements in reasoning through methods that encourage structured intermediate steps. CoT prompting enables models to generate interpretable step-by-step reasoning traces and improves accuracy across complex reasoning tasks \cite{wei2022cot,kojima2022zeroshot,sprague2025cotcotchainofthoughthelps}. To enhance reliability and mitigate stochastic failures, self-consistency samples multiple reasoning paths and selects the answer that is most frequently generated \cite{wang2023selfconsistency}. Other prominent techniques include question decomposition and planning-based frameworks, such as Least-to-Most prompting and Tree-of-Thoughts, which break complex queries into subtasks or explore alternative reasoning trajectories to strengthen coherence and coverage \cite{zhou2022leasttomost,yao2023tree}. Recent advances have further expanded these capabilities through test-time compute scaling, where models allocate additional inference computation to improve reasoning performance \cite{snell2025scaling,bi2025forest,muennighoff2025s1,zhang2025testtimescaling}. Self-refinement methods enable models to iteratively critique and improve their own outputs without additional training \cite{madaan2023selfrefine}, while process reward models provide fine-grained step-level supervision to guide reasoning and correct intermediate errors \cite{setlur2025rewarding,zhang2025lessons,khalifa2025thinkprm}. Additionally, tool-augmented approaches extend intrinsic reasoning capabilities by integrating external retrieval or code execution modules, facilitating grounded or verifiable computation, as exemplified by ReAct, Toolformer, and Program-of-Thoughts prompting \cite{yao2023react,schick2023toolformer,chen2023pot}. While these methods often improve reasoning performance, they also introduce additional inference overhead through longer generations, tool calls, or multiple reasoning paths, motivating methods that preserve reasoning quality while reducing unnecessary computation \cite{samsi2023wordstowatts,fernandez2025energy,wu2025inference}.

\subsection{Uncertainty Estimation in Reasoning}
Uncertainty estimation has become an essential diagnostic tool for assessing and improving the quality of reasoning in LLMs. Prior work has examined how model confidence, often derived from token probabilities, correlates with correctness and can reveal whether a model can distinguish reliable answers from unreliable ones \cite{kadavath2022lmknow}. Other studies have investigated whether LLMs can faithfully verbalize their intrinsic uncertainty or learn to express uncertainty through self-training \cite{yona2024faithfully,liu2024llmslearnuncertainty}. Beyond single-path diagnostics, semantic-based approaches measure the variability across semantically equivalent generations to estimate epistemic uncertainty and detect hallucinations \cite{kuhn2023semantic,baan2023uncertainty,Farquhar2024,huang2025surveyhallucination}. More recent work studies uncertainty directly over reasoning traces, showing that CoT can improve response-wise uncertainty quantification and that internal reasoning states can support efficient test-time reasoning assessment \cite{zhang2025cot,ni2025reprobe}. Confidence-signal enhanced reasoning assigns confidence scores to intermediate reasoning steps and uses them to weight and aggregate multiple CoT paths, improving robustness over vanilla self-consistency by explicitly leveraging uncertainty structure \cite{razghandi2025cerconfidenceenhancedreasoning}. Unlike prior work that primarily uses uncertainty for calibration, hallucination detection, or reweighting multiple completed reasoning paths, our approach uses uncertainty-derived features from a single trajectory to determine whether further sampling is necessary, thereby reducing the cost of multi-path reasoning.

\subsection{Adaptive and Early-Exit Reasoning}
To mitigate the high computational burden associated with extended reasoning traces, adaptive test-time computation aims to allocate inference resources according to question difficulty rather than applying the same computation budget to every question \cite{snell2025scaling,wu2025inference,alomrani2025reasoning}. In multi-path reasoning, dynamic voting mitigates inconsistency by aggregating predictions from multiple reasoning trajectories and terminating sampling once sufficient agreement is observed \cite{xue2023dynamic}. More recent adaptive self-consistency methods go beyond count-based voting by incorporating rationale quality or response-level reliability signals to determine whether additional samples are needed \cite{wan2025reasoning,kim2026reliability}. However, these strategies 
often require sampling multiple complete reasoning chains or observing response-level agreement before deciding to stop, which remains costly in terms of token usage and computational overhead.

Another line of work trains a decision model to learn dynamic stopping or computation-allocation policies. Confident adaptive language modeling introduces token-level early exits inside the Transformer and calibrates their confidence so that the final sequence quality remains within a user-specified tolerance of full computation \cite{schuster2022calm}. Earlier adaptive computation methods similarly learn how much internal computation to use for each input, either by varying the number of recurrent update steps or by exiting at different network depths \cite{graves2016act,teerapittayanon2016branchynet,xIN2020deebert}. Recent latent-reasoning approaches train models to decide when to stop internal reasoning, often using reinforcement learning or latent-state objectives \cite{hao2024training,ning2025learning}. Related compute-allocation work formulates test-time computation as a constrained optimization problem and trains lightweight policies to allocate additional compute only to difficult instances \cite{zhai2026adaptive}. However, these stopping or compute-allocation policies are often tied to architecture-specific internal representations, latent-state training objectives, or model-specific calibration, which may limit transfer across LLM backbones, prompt formats, and benchmark domains.

\begin{figure*}[t]
    \centering
    \includegraphics[width=1.0\linewidth] {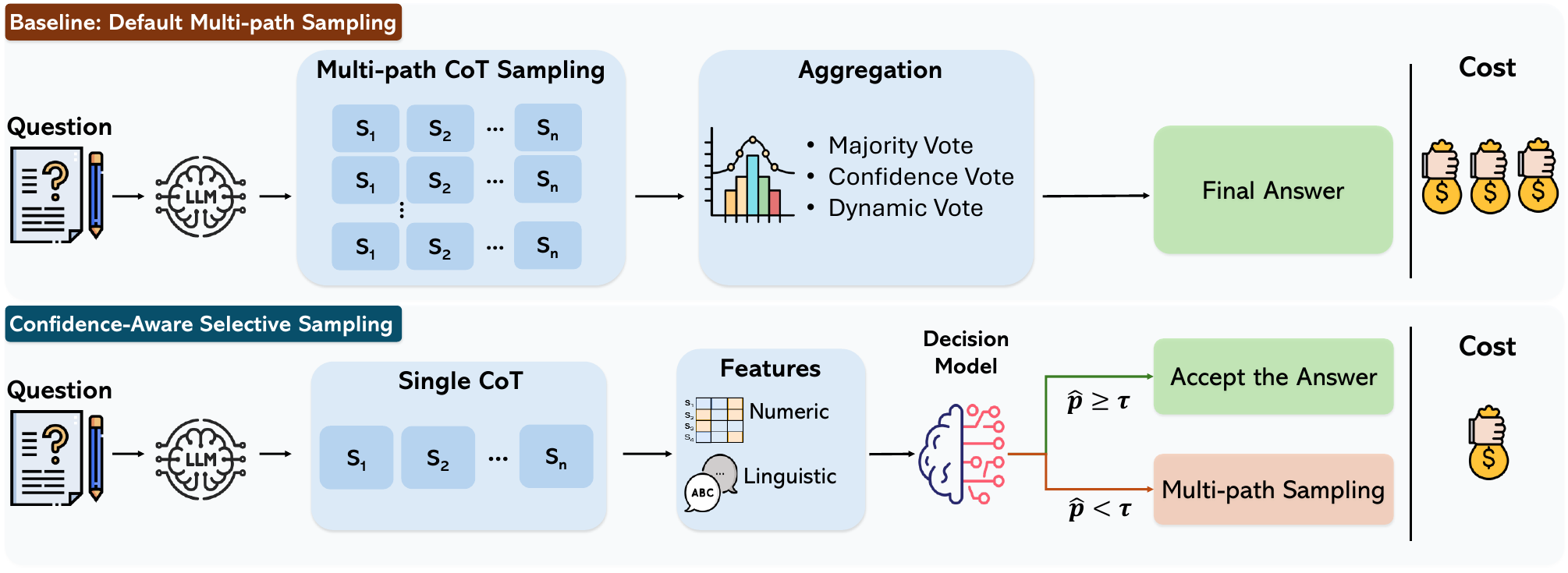}
    \caption{
    \textbf{Overview of the framework.} For each question, an LLM generates a reasoning trajectory, from which numeric and linguistic features are extracted. Our decision model then analyzes this sequence of features and estimates the probability $\hat p$ that the initial answer will be correct or wrong. A confidence threshold $\tau$ determines whether additional multi-path reasoning is necessary. Unlike the default multi-path sampling method, our confidence-aware selective sampling reduces inference cost. 
    }

    \label{fig:early_exit_framework}
\end{figure*}

\section{Method}


\subsection{Sentence-Level Per-Choice Prediction from Logits}

To monitor the reasoning dynamics during CoT generation, we derive sentence-level answer-choice probabilities from token-level logits. Given a question $q$ with $K$ answer choices $\{A_i\}_{i=1}^{K}$, an LLM generates a CoT reasoning trace $r$ token by token. We segment the generated reasoning trace into sentences $\{s_n\}_{n=1}^{N}$ and estimate the model's belief over answer choices after each sentence.

Let $r_{1:n}$ denote the reasoning prefix up to the $n$-th sentence, i.e.,
$r_{1:n}=s_1\oplus\cdots\oplus s_n$. To estimate the likelihood of each option $A_i$ given this prefix, we compute the conditional log-probability score:
\begin{equation}
    \alpha_{n,i}
    =
    \frac{1}{|A_i|}
    \sum_{j=1}^{|A_i|}
      \log p(a_{ij} \mid q, r_{1:n}, a_{i,<j}),
\end{equation}
where $a_{ij}$ is the $j$-th token of answer option $A_i$. We use length normalization to reduce bias toward shorter answer options. We then normalize these scores across all options:
\begin{equation}
    p_n(A_i)
    =
    \frac{\exp(\alpha_{n,i})}
    {\sum_{k=1}^{K}\exp(\alpha_{n,k})}.
\end{equation}

This yields a probability vector 
$\mathbf{p}_n = [p_n(A_1), p_n(A_2), \dots, p_n(A_K)]$
for each $s_n$, representing the model's evolving 
beliefs over answer choices. These per-sentence trajectories are stored for later analysis and used to construct the feature trajectory for the selective-sampling stage.


\subsection{Decision Model}

Given a question, the LLM produces a completed CoT trajectory and a final answer. A GRU-based decision model estimates the probability that the final answer is correct based on the completed trajectory. The decision model outputs a scalar probability $\hat{p}\in[0,1]$ and applies a routing threshold $\tau$:
\begin{equation}
\begin{cases}
\text{likely-correct,} & \text{if } \hat{p} \ge \tau,\\
\text{likely-wrong,}   & \text{otherwise}.
\end{cases}
\label{eq:triage_rule}
\end{equation}

Predictions in the likely-correct group accept the final answer, while predictions in the likely-wrong group are routed to the multi-path aggregation module, instantiated as dynamic voting in our experiments. 
\subsubsection{Feature Extraction}
\label{sec:feature_extraction}

We extract a sentence-level trajectory from the completed CoT. The CoT text is segmented into $N$ sentences $\{s_n\}_{n=1}^{N}$ using punctuation and newline boundaries. For each sentence $s_n$, we retain a scalar confidence signal $p_n\in(0,1)$ derived from the answer-choice distribution, an uncertainty signal $H_n$ computed from token-level generation entropy, and the prefix length $L_n$, defined as the cumulative number of word tokens up to sentence $n$. The resulting trajectory is
\begin{equation}
\mathcal{Z}=\{(p_n,H_n,L_n,s_n)\}_{n=1}^{N}.
\label{eq:traj}
\end{equation}
We then construct $\mathbf{X}\in\mathbb{R}^{N\times D}$ by concatenating trajectory-based numeric features with sentence-level linguistic features, yielding $D=32$ features per sentence. Additional preprocessing and implementation details are provided in Appendix~\ref{appendix:implementation}.

\paragraph{Numeric trajectory features.}
Table~\ref{tab:num_features} summarizes the numeric features computed from $(p_n,H_n,L_n)$. These features capture local confidence and uncertainty, prefix-level information, temporal changes, and short-term stability along the completed reasoning trajectory.

\begin{table}[t]
\centering
\small
\caption{Numeric trajectory features computed at sentence position $s_n$.}
\vspace{-2mm}
\setlength{\tabcolsep}{4pt}
\begin{tabularx}{\linewidth}{c l X}
\hline
Idx & Feature & Description \\
\hline
1  & $p_n$ & Scalar confidence signal derived from the answer-choice distribution after sentence $s_n$. \\
2  & $H_n$ & Mean next-token generation entropy over the tokens in sentence $s_n$. \\
3  & $p_n/\log(1+L_n)$ & Confidence normalized by the cumulative prefix length. \\
4  & $\Delta p_n$ & $p_n-p_{n-1}$ \\
5  & $\Delta H_n$ & $H_n-H_{n-1}$. \\
6  & $\sigma^{(k)}_n$ & Standard deviation over the last up to $k$ confidence values. \\
7  & $r^{(k)}_n$ & Range over the last up to $k$ confidence values. \\
8  & $L_n$ & Cumulative number of word tokens up to and including sentence $s_n$. \\
9  & $\mathrm{EMA}(p)_n$ & Exponential moving average of the confidence signal up to sentence $n$. \\
10 & $\Delta\mathrm{EMA}(p)_n$ & $\mathrm{EMA}(p)_n-\mathrm{EMA}(p)_{n-1}$. \\
11 & $z(p_n)$ & Trajectory-level z-score of the confidence signal. \\
12 & $z(\mathrm{EMA}(p)_n)$ & Trajectory-level z-score of the EMA confidence signal. \\
\hline
\end{tabularx}

\label{tab:num_features}
\end{table}

\paragraph{Linguistic features.}
In addition to numeric signals, we extract lightweight linguistic features from the sentence text $s_n$ and its relation to the prompt (question and answer options). These features capture text statistics, reasoning style, and topical alignment, without using any text embeddings. Table~\ref{tab:ling_features} summarizes the linguistic features.

\begin{table}[t]
\centering
\small
\caption{Sentence-level linguistic features extracted from each sentence $s_n$.}
\vspace{-2mm}
\setlength{\tabcolsep}{4pt}
\begin{tabularx}{\linewidth}{l X}
\hline
Category & Features \\
\hline
Sentence statistics &
token count; character count; average token length of $s_n$ \\

Lexical composition &
stop word ratio; content word ratio in $s_n$ \\

Punctuation &
comma, period, question, and exclamation counts; punctuation density in $s_n$ \\

Character patterns &
digit ratio; uppercase ratio in $s_n$ \\

Prompt overlap &
overlap counts between $s_n$ and the question/options; normalized overlap ratios \\

Reasoning markers &
hedge word count; certainty word count; logical connector count in $s_n$ \\

Position &
normalized position of $s_n$ within the CoT \\
\hline
\end{tabularx}

\label{tab:ling_features}
\end{table}

\subsubsection{Model Architecture}
\label{sec:model_architecture}

We instantiate the trainable component as a GRU-based decision model that operates on the complete feature trajectory $\mathbf{X}\in\mathbb{R}^{N\times D}$ extracted from a completed CoT. The model consists of the following blocks.

\paragraph{Feature gating block.}
We employ a trajectory-conditioned feature gating block to adaptively reweight the input feature sequence $\mathbf{X}$. We first compute a trajectory-level summary by mean pooling over valid sentence positions:
\[
\mathbf{s}
=
\frac{\sum_{n=1}^{N} m_n \mathbf{X}_n}
{\sum_{n=1}^{N} m_n},
\]
where $m_n\in\{0,1\}$ indicates whether sentence position $n$ is valid. The summary vector is passed through a lightweight two-layer multilayer perceptron (MLP) with ReLU and sigmoid activations to produce channel-wise gating weights
$\mathbf{g}\in(0,1)^D$. The gated feature sequence is then computed as
\[
\tilde{\mathbf{X}}_n=\mathbf{X}_n\odot \mathbf{g}.
\]
The same gating vector is applied to all sentence positions within a trajectory, allowing the decision model to emphasize or suppress feature dimensions according to the global characteristics of the completed reasoning trajectory.

\paragraph{GRU encoder block.}
The gated sequence $\tilde{\mathbf{X}}$ is encoded by a single-layer unidirectional gated recurrent unit (GRU) with hidden size $d_h=64$. The GRU produces contextual representations
\[
\mathbf{U}=[\mathbf{u}_1,\ldots,\mathbf{u}_N]\in\mathbb{R}^{N\times d_h},
\]
which capture temporal dependencies in the feature dynamics across the reasoning trajectory. For variable-length trajectories, padded sentence positions are excluded from recurrent encoding using packed sequence representations.

\paragraph{Multi-head self-attention block.}
The sequence of contextual representations $\mathbf{U}$ is further processed by a lightweight 4-head self-attention block. 
The block follows a pre-normalization transformer-style structure consisting of LayerNorm, multi-head self-attention with residual connections, and a position-wise feed-forward sublayer. This component allows each sentence representation to incorporate information from other reasoning steps, enhancing the modeling of long-range dependencies across the trajectory.

\paragraph{Position-wise projection head.}
Each contextual representation is mapped to a scalar logit by a position-wise MLP. The head consists of LayerNorm followed by a two-layer MLP with hidden dimension 32 and ReLU activation, producing per-sentence logits
$\{\ell_n\}_{n=1}^{N}$ and probabilities
\[
q_n=\operatorname{sigmoid}(\ell_n).
\]
The final-position probability $\hat{p}=q_N$ is used as the trajectory-level score for detecting whether the final answer is likely to be correct.


\subsubsection{Threshold Calibration}

We determine dataset-specific confidence thresholds using validation-set profiling. Specifically, we sweep the confidence threshold $\tau \in [0,1]$ and evaluate both prediction accuracy and token reduction on the validation set. For each dataset, we first identify the maximum validation accuracy achieved across all candidate thresholds. We then select the threshold that yields the largest token reduction while keeping the validation accuracy within 0.5 percentage points of this maximum. This procedure prioritizes efficiency gains while preventing material degradation in predictive performance.

\begin{figure*}[t]
    \centering
    \includegraphics[width=1.0\linewidth]{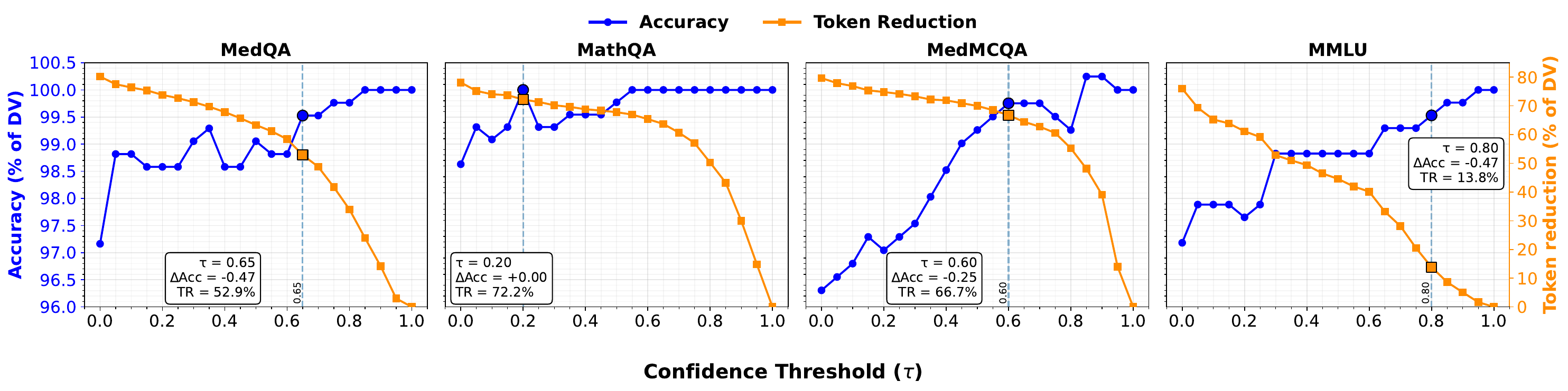}
    \caption{ \textbf{Confidence Threshold Calibration.} Accuracy (blue, \% of DV) and token reduction (orange, \% vs.\ DV) versus confidence threshold $\tau$ on MedQA, MathQA, MedMCQA, and MMLU. $\tau=1.0$ is the DV baseline (100\% accuracy, 0\% reduction). 
    }
    \label{fig:tradeoff-medqa-gptoss20b}
\end{figure*}

\section{Experiments}

\paragraph{Datasets}

We evaluate our method on four multiple-choice question answering datasets that cover both medical and general domains.
\begin{itemize}
\item \textbf{MedQA} \cite{jin2021medqa}:  
A dataset derived from medical licensing exam questions, containing expert-authored problems across multiple clinical topics. It assesses medical reasoning and factual recall in clinical question-answering settings.

\item \textbf{MedMCQA} \cite{pal2022medmcqa}:  
A challenging corpus of medical questions collected from Indian medical entrance exams. It contains over 200{,}000 questions and tests domain comprehension and knowledge integration.

\item \textbf{MathQA} \cite{amini2019mathqa}: 
A dataset of arithmetic and commonsense mathematical problems requiring multi-sentence symbolic reasoning. We use the multiple-choice version standardized by prior reasoning benchmarks.

\item \textbf{MMLU} \cite{hendrycks2020measuring}: 
A broad evaluation benchmark spanning 57 subjects, ranging from science, technology, engineering, and mathematics (STEM) to humanities, designed to assess general knowledge and reasoning in large language models.
\end{itemize}

\paragraph{Models:} We evaluate our framework on a diverse set of open-source LLMs spanning multiple architectures and parameter scales. The experiments are conducted using GPT-OSS 20B \cite{agarwal2025gpt}, Meta Llama 3.1 8B Instruct \cite{dubey2024llama}, Qwen 2.5 7B \cite{qwen2025qwen25technicalreport}, and Qwen 3 14B/32B \cite{yang2025qwen3technicalreport}. All of the models support autoregressive CoT generation with probability outputs, enabling a unified routing pipeline across different reasoning engines. We report evaluation results for GPT-OSS 20B in the main experiments and defer results for the other models to Appendix~\ref{appendix:a} and Appendix~\ref{appendix:b}.

\paragraph{Baselines:} We compare our approach against representative multi-path reasoning baselines and confidence-aware variants:
\begin{itemize}
    \item \textbf{Self-Consistency (SC):} Aggregates multiple sampled reasoning paths and selects the most frequent answer to enhance reasoning accuracy \cite{wang2023selfconsistency}.
    \item \textbf{Confidence Enhanced Reasoning (CER):} Aggregates multiple reasoning paths by weighting them according to confidence estimated from reasoning steps \cite{razghandi2025cerconfidenceenhancedreasoning}.
    \item \textbf{Dynamic Voting (DV):} Adaptively samples reasoning paths and stops once a voting consensus is reached, reducing the computation relative to self-consistency \cite{xue2023dynamic}.
\end{itemize}

\paragraph{Experiment Setup:} We train and evaluate all decision models using the same data splits on a single NVIDIA H100 GPU. For each LLM, the decision model is trained only on the MedQA training split. For each target dataset, we use 500 validation examples solely for threshold calibration and 1,000 held-out examples for testing. No decision-model parameters are updated outside MedQA. 
We sample 10 CoTs for multi-path experiments with a temperature of 1.0. 
Full data preprocessing and training details are provided in Appendix~\ref{appendix:implementation}.

\section{Results}


\begin{figure*}[t]
    \centering
    \includegraphics[width=.7\linewidth]{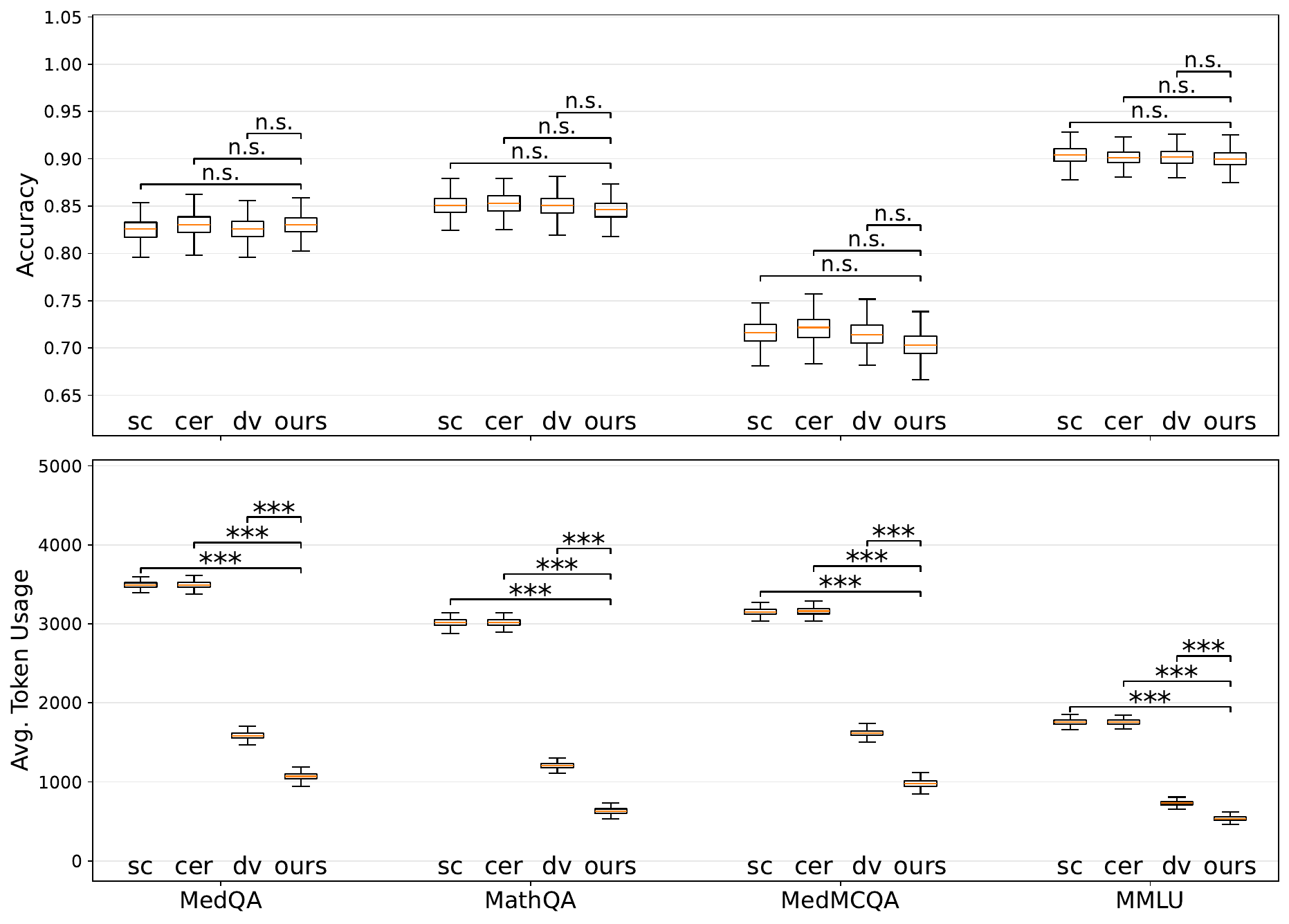}
    \caption{\textbf{Distribution of accuracy and token usage across datasets.}
Boxplots show the performance of SC, CER, DV, and Ours on MedQA, MathQA, MedMCQA, and MMLU using GPT-OSS 20B. The top panel reports accuracy, while the bottom panel reports average token usage. Statistical significance is evaluated using paired bootstrap with 2{,}000 resamples; 
$n.s.$ indicates not significant and $^{*}$ indicates $p<0.05$; $^{**}$ indicates $p<0.01$; $^{***}$ indicates $p<0.001$. Higher accuracy is better, whereas lower token usage indicates greater efficiency.}
    \label{fig:gpt_boxplot}
\end{figure*}

\subsection{Confidence Threshold Calibration}

Figure~\ref{fig:tradeoff-medqa-gptoss20b} illustrates the accuracy (blue) versus token reduction (orange) tradeoffs across confidence threshold choices.  
The selected optimal thresholds correspond to the maximum token reduction under the $\leq 0.5\%$ accuracy-drop constraint. 
Based on this profiling, we selected: $\tau_{\text{MedQA}} = 0.65$, $\tau_{\text{MathQA}} = 0.20$, $\tau_{\text{MedMCQA}} = 0.60$, and $\tau_{\text{MMLU}} = 0.80$. The calibration results for other LLMs are shown in Appendix~\ref{appendix:a} Figures~\ref{fig:llama_cali}--\ref{fig:qwen3-32_cali}.

\subsection{Accuracy and Efficiency}

Figure~\ref{fig:gpt_boxplot} presents a comparison of accuracy and token usage between our proposed model and the baselines for GPT-OSS 20B. 
Across all four datasets, the differences in accuracy between our method and the multi-path baselines were not statistically significant ($n.s.$) under bootstrap testing. In MedQA, MathQA, MedMCQA and MMLU, our performance remains within a narrow margin of SC, CER, and DV, indicating that selective multi-path reasoning preserves the final prediction accuracy.

In contrast, the token usage reductions of our framework are substantial and statistically significant ($p<0.05$) when comparing to SC, CER, and DV on every dataset. Relative to SC and CER, our framework reduces token consumption by approximately 69--79\% across tasks. Even compared to DV, token consumption is reduced by 27--48\%. 
The tradeoff results for remaining LLMs are shown in Appendix~\ref{appendix:b}, Figures~\ref{fig:llama_boxplot}--\ref{fig:qwen3-32_boxplot}, which show the same finding: our solution significantly reduces token usage while accuracy remains close to multi-path baselines in most settings.

For each LLM, we train the decision model on MedQA CoT traces and evaluate it on MedQA, as well as under calibration-only transfer to MathQA, MedMCQA, and MMLU. For all datasets, only the routing threshold is calibrated on the corresponding validation set, while the decision-model parameters are kept fixed. Since different LLMs exhibit distinct reasoning dynamics, we train a separate decision model for each LLM while keeping the feature extraction pipeline and model architecture fixed.

As shown for GPT-OSS 20B in Figure~\ref{fig:gpt_boxplot}, our method achieves accuracy comparable to multi-path reasoning while substantially reducing token usage. Similar trends are observed for other models, as shown in Appendix~\ref{appendix:b}, Figure~\ref{fig:llama_boxplot}--\ref{fig:qwen3-32_boxplot}. These suggest that the decision model can learn a trajectory-based decision rule that transfers across domains.

We further observe that larger models, such as GPT-OSS 20B, tend to produce more separable confidence and uncertainty trajectories between correct and incorrect reasoning paths. This clearer separation allows the decision model to accept initial reasoning more frequently, leading to larger efficiency gains. In contrast, smaller models, such as Llama 3.1 8B Instruct, produce noisier trajectory signals, which limits token savings but still preserves the overall accuracy--efficiency advantage of the proposed approach.



\subsection{Ablation Study}

\begin{table}[t]
\centering
\footnotesize
\setlength{\tabcolsep}{4pt}
\caption{\textbf{Ablation of Feature Gating (FG) block and Multi-Head Self-Attention (MHSA) block}. The variant without FG and MHSA serves as the reference. Best results are in bold. $\Uparrow$: higher is better; $\Downarrow$: lower is better.}
\renewcommand{\arraystretch}{0.95}
\begin{adjustbox}{width=\linewidth}
\begin{tabular}{llll}
\toprule
Dataset & Variant & Accuracy$\Uparrow$ & Token Usage$\Downarrow$  \\
\midrule

\multirow{4}{*}{MedQA}
& FG \xmark, MHSA \xmark & 0.822 & 1230 \\
& FG \cmark, MHSA \xmark & 0.825\up{0.003} & 1185\down{3.66\%} \\
& FG \xmark, MHSA \cmark & 0.828\up{0.006} & 1206\down{1.95\%} \\
& \textbf{FG \cmark, MHSA \cmark} & \textbf{0.830\up{0.008}} & \textbf{1026\down{16.59\%}} \\

\midrule
\multirow{4}{*}{MathQA}
& FG \xmark, MHSA \xmark & 0.847 & 704 \\
& FG \cmark, MHSA \xmark & \textbf{0.848\up{0.001}} & 689\down{2.13\%} \\
& FG \xmark, MHSA \cmark & 0.847\up{0.000} & 692\down{1.70\%} \\
& \textbf{FG \cmark, MHSA \cmark} & 0.847\up{0.000} & \textbf{629\down{10.65\%}} \\

\midrule
\multirow{4}{*}{MedMCQA}
& FG \xmark, MHSA \xmark & \textbf{0.704} & 1164 \\
& FG \cmark, MHSA \xmark & 0.701\down{0.003} & 1100\down{5.50\%} \\
& FG \xmark, MHSA \cmark & 0.701\down{0.003} & 1133\down{2.66\%} \\
& \textbf{FG \cmark, MHSA \cmark} & \textbf{0.704\up{0.000}} & \textbf{983\down{15.55\%}} \\

\midrule
\multirow{4}{*}{MMLU}
& FG \xmark, MHSA \xmark & \textbf{0.900} & 691 \\
& FG \cmark, MHSA \xmark & \textbf{0.900\up{0.000}} & 576\down{16.64\%} \\
& FG \xmark, MHSA \cmark & 0.898\down{0.002} & 570\down{17.51\%} \\
& \textbf{FG \cmark, MHSA \cmark} & \textbf{0.900\up{0.000}} & \textbf{536\down{22.43\%}} \\

\midrule
\multicolumn{4}{c}{\textit{Average Performance Across All Datasets}} \\
\midrule
& FG \xmark, MHSA \xmark & 0.818 & 947 \\
& FG \cmark, MHSA \xmark & 0.819\up{0.001} & 888\down{6.23\%} \\
& FG \xmark, MHSA \cmark & 0.819\up{0.001} & 900\down{4.96\%} \\
& \textbf{FG \cmark, MHSA \cmark} & \textbf{0.820\up{0.002}} & \textbf{794\down{16.16\%}} \\

\bottomrule
\end{tabular}
\end{adjustbox}
\label{tab:ablation}
\end{table}

\begin{table}[t]
\centering
\footnotesize
\setlength{\tabcolsep}{4pt}
\caption{\textbf{Feature Ablation study.} We compare training with numeric (reference), linguistic, and both feature types combined.  
$\Uparrow$: higher is better; $\Downarrow$: lower is better.}
\renewcommand{\arraystretch}{0.95}
\begin{adjustbox}{width=\linewidth}
\begin{tabular}{llll}
\toprule
Dataset & Variant & Accuracy$\Uparrow$ & Token Usage$\Downarrow$ \\
\midrule

\multirow{3}{*}{MedQA}
& Numeric only              & 0.825  & 1148 \\
& Linguistic only           & 0.825\up{0.000}  & 1222\up{6.45\%} \\
& \textbf{Numeric + Linguistic} & \textbf{0.830\up{0.005}} & \textbf{1070\down{6.79\%}} \\
\midrule

\multirow{3}{*}{MathQA}
& Numeric only              & 0.845  & 736 \\
& Linguistic only           & 0.847\up{0.002}  & 681\down{7.47\%} \\
& \textbf{Numeric + Linguistic} & \textbf{0.847\up{0.002}} & \textbf{630\down{14.40\%}} \\
\midrule

\multirow{3}{*}{MedMCQA}
& Numeric only              & 0.704  & 1069 \\
& Linguistic only           & 0.702\down{0.002} & 1144\up{7.02\%} \\
& \textbf{Numeric + Linguistic} & \textbf{0.704\up{0.000}} & \textbf{983\down{8.04\%}} \\
\midrule

\multirow{3}{*}{MMLU}
& Numeric only              & 0.898  & 583 \\
& Linguistic only           & 0.899\up{0.001}  & 595\up{2.06\%} \\
& \textbf{Numeric + Linguistic} & \textbf{0.900\up{0.002}} & \textbf{536\down{8.06\%}} \\
\midrule

\multicolumn{4}{c}{\textit{Average Performance Across All Datasets}} \\
\midrule
& Numeric only              & 0.818  & 884 \\
& Linguistic only           & 0.818\up{0.000}  & 911\up{3.05\%} \\
& \textbf{Numeric + Linguistic} & \textbf{0.820\up{0.002}} & \textbf{805\down{8.94\%}} \\

\bottomrule
\end{tabular}
\end{adjustbox}
\label{tab:feature_ablation}
\end{table}

We conduct two ablation studies to analyze the contributions of key architectural design choices and input feature types.

We first evaluate the contribution of the feature gating (FG) block and the multi-head self-attention (MHSA) block (Table~\ref{tab:ablation}). Using either module alone yields only limited or inconsistent improvements over the variant with both blocks disabled. In contrast, enabling both modules consistently achieves the best accuracy--efficiency trade-off. 

Similarly, to determine the optimal feature set, we compare numeric-only, linguistic-only, and combined features (Table~\ref{tab:feature_ablation}). 
Combining the two yields the best performance, improving both accuracy and token efficiency.

\section{Discussion and Conclusion}


In this paper, we proposed a confidence-aware selective sampling framework that improves the accuracy--efficiency trade-off of LLM reasoning. Rather than applying costly multi-path reasoning to every query, our framework analyzes the sentence-level dynamics of a completed reasoning trajectory and predicts whether additional reasoning is needed. Empirically, we find that sentence-level features, including probability trends, entropy dynamics, and convergence patterns, provide effective signals of reasoning reliability. We further show that these signals transfer across datasets and domains: a decision model trained on MedQA traces transfers to MathQA, MedMCQA, and MMLU with only threshold calibration. Finally, trajectories from larger LLMs exhibit clearer separability between reliable and unreliable reasoning paths, suggesting that improving trajectory quality and uncertainty separability remains an important direction for smaller LLMs.


\section{Limitations}

Our work has several limitations that can serve as a basis for further investigation. First, the framework is evaluated primarily on multiple-choice scientific and medical question answering tasks, where reasoning trajectories exhibit relatively structured patterns. Its effectiveness for open-ended generation, long-form reasoning, or dialogue settings can be explored in future research. Second, the method analyzes completed reasoning trajectories and therefore cannot be directly used for online early-exit decisions during generation. Adapting the framework to operate causally in intermediate sentences would require additional modeling and calibration in future work. Third, monitoring numeric and linguistic features depends on access to internal signals and is mainly validated on open-source LLMs. Future work could improve the proposed method by focusing solely on the reasoning trajectory to evaluate proprietary LLMs.


\bibliography{custom}

\appendix

\section{Implementation Details}
\label{appendix:implementation}

\paragraph{CoT generation and answer extraction.}
For each question, we construct a chat-style prompt containing the question and answer options, and instruct the LLM to produce concise step-by-step reasoning followed by a final answer line in the format \texttt{Answer: X}, where \texttt{X} is one of the provided option letters. The CoT portion is separated from the final answer line using this marker. If the final answer marker is absent, we fall back to forced-choice scoring over the candidate answer letters using the completed CoT prefix.

\paragraph{Sentence-level probability and entropy extraction.}
For each sentence prefix $r_{1:n}$, we append an answer prompt and compute length-normalized conditional log-probabilities for all candidate answer options. The resulting scores are normalized with a softmax to obtain the answer-choice distribution. Token-level generation entropy is computed from the next-token distribution during CoT generation and then averaged over the tokens aligned to each sentence span. Sentence spans are obtained using punctuation-based segmentation, and token-to-sentence alignment is performed using tokenizer offset mappings.

\paragraph{Trajectory preprocessing.}
For each trajectory, we keep only sentence positions where both the confidence signal and entropy signal are available and finite. The prefix length $L_n$ is computed as the cumulative number of regex-based word tokens up to sentence $n$. Rolling statistics use a window of up to $k=4$ recent sentence-level confidence values. The exponential moving average uses smoothing coefficient $\alpha=0.8$. Trajectory-level z-scores are computed within each completed CoT trajectory.

\paragraph{Linguistic feature extraction.}
The linguistic feature set is rule-based and does not use text embeddings. Stop words are removed when computing content-word overlap with the question and answer options. Hedge, certainty, and logical-marker counts are computed using fixed lexicons. Hedge words include terms such as \textit{maybe}, \textit{might}, \textit{could}, \textit{possibly}, \textit{perhaps}, \textit{appears}, \textit{seems}, \textit{suggests}, \textit{likely}, and \textit{unlikely}. Certainty words include \textit{therefore}, \textit{thus}, \textit{hence}, \textit{certainly}, \textit{clearly}, \textit{definitely}, \textit{must}, \textit{always}, and \textit{never}. Logical markers include \textit{because}, \textit{so}, \textit{therefore}, \textit{thus}, \textit{hence}, \textit{however}, \textit{but}, \textit{although}, \textit{while}, and \textit{whereas}.

\paragraph{Training objective.}
The decision model is trained as a binary classifier. The label is $y=1$ if the initial CoT answer is correct and $y=0$ otherwise. We use binary cross-entropy loss with masking over valid sentence positions for variable-length trajectories. To account for class imbalance, the positive class weight is set to the ratio between the number of negative and positive training examples.

\paragraph{Optimization and model selection.}
We train the decision model with AdamW using a learning rate of $2\times10^{-3}$, weight decay of $10^{-4}$, batch size 64, and 20 training epochs. Gradients are clipped to a maximum norm of 1.0. We use random seed 1337 for Python, NumPy, and PyTorch. The checkpoint with the lowest validation loss is selected for evaluation.

\paragraph{Sequence batching and padding.}
CoT trajectories have variable numbers of sentences. During batching, trajectories are padded to the maximum sequence length within each batch, and padded sentence positions are masked during training and evaluation. No fixed sentence-level maximum length is imposed beyond the generation and preprocessing limits.

\paragraph{Reproducibility details.}
We use the tokenizer chat template when available and otherwise fall back to a plain role-based prompt. For multi-path baselines, we sample up to 10 reasoning trajectories with temperature 1.0 using the same prompt and answer-extraction procedure. For each trajectory, we extracted CoT, answer-choice probabilities, sentence-level entropy values, and number of generated tokens.

\section{Confidence Threshold Calibration for Additional LLMs}
\label{appendix:a}

\begin{figure*}[!t]     
  \centering
  \includegraphics[width=1.0\linewidth]{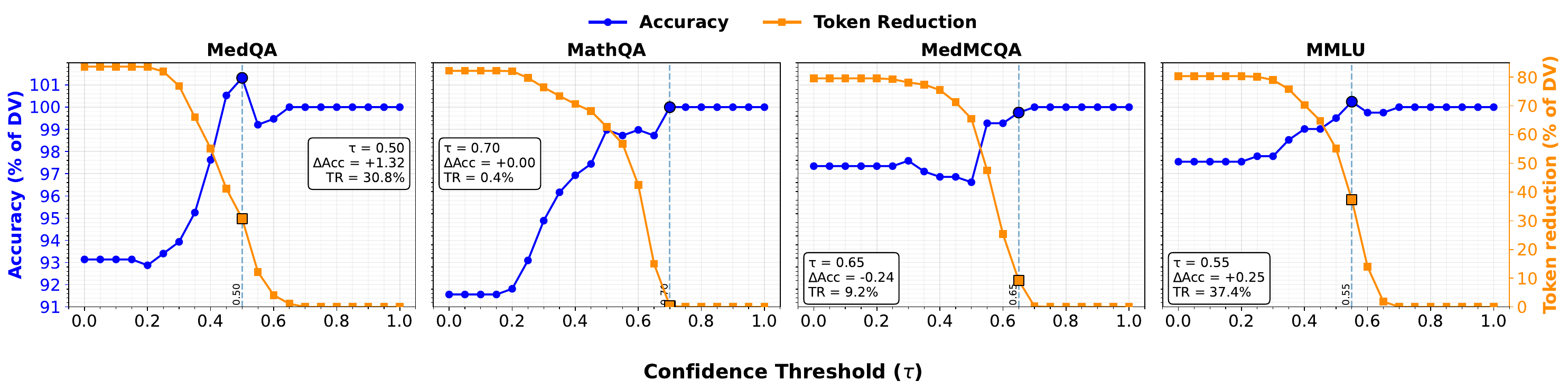}
  \caption{\textbf{Confidence Threshold Calibration for Llama 3.1 8B Instruct.} Accuracy (blue, \% of DV) and token reduction (orange, \% vs.\ DV) versus confidence threshold $\tau$ on MedQA, MathQA, MedMCQA, and MMLU. $\tau=1.0$ is the DV baseline (100\% accuracy, 0\% reduction).}
  \label{fig:llama_cali}
\end{figure*}

\begin{figure*}[t]
    \centering
    \includegraphics[width=1.0\linewidth]{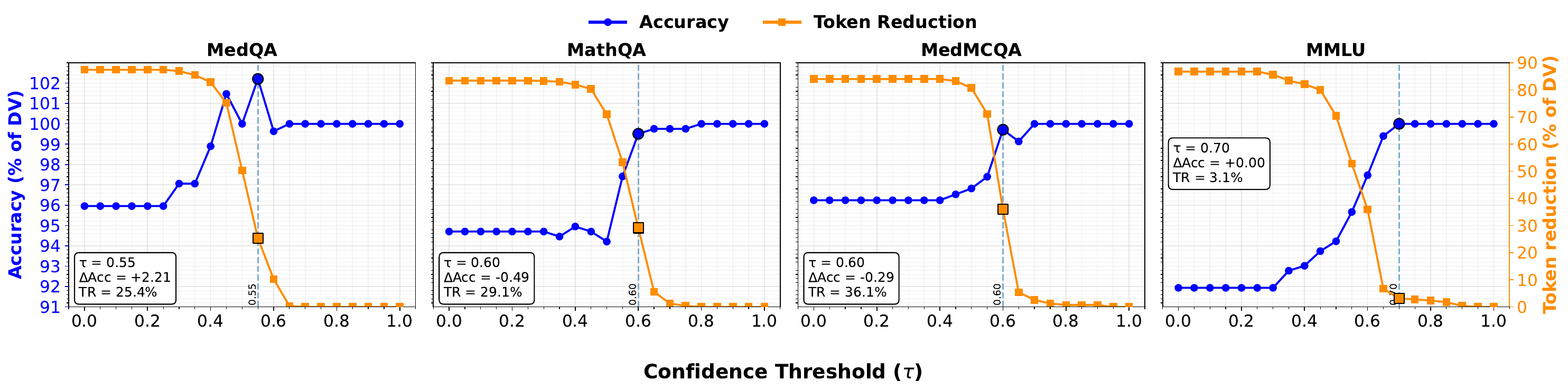}
    \caption{ \textbf{Confidence Threshold Calibration for Qwen 2.5 7B.} Accuracy (blue, \% of DV) and token reduction (orange, \% vs.\ DV) versus confidence threshold $\tau$ on MedQA, MathQA, MedMCQA, and MMLU. $\tau=1.0$ is the DV baseline (100\% accuracy, 0\% reduction).
    }
    \label{fig:qwen2_cali}
\end{figure*}

\begin{figure*}[t]
    \centering
    \includegraphics[width=1.0\linewidth]{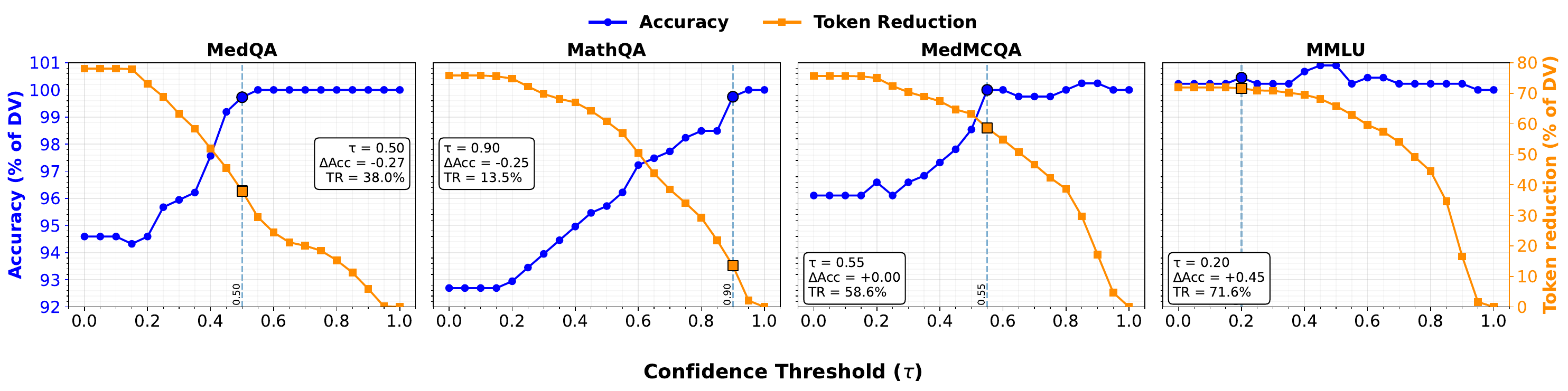}
    \caption{ \textbf{Confidence Threshold Calibration for Qwen 3 14B.} Accuracy (blue, \% of DV) and token reduction (orange, \% vs.\ DV) versus confidence threshold $\tau$ on MedQA, MathQA, MedMCQA, and MMLU. $\tau=1.0$ is the DV baseline (100\% accuracy, 0\% reduction).
    }
    \label{fig:qwen3-14_cali}
\end{figure*}

\begin{figure*}[t]
    \centering
    \includegraphics[width=1.0\linewidth]{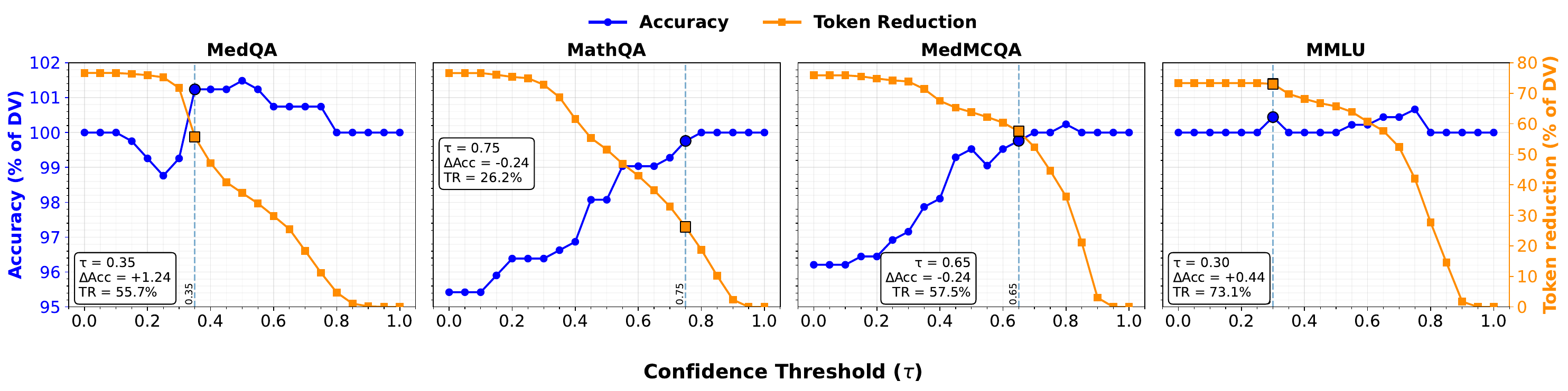}
    \caption{ \textbf{Confidence Threshold Calibration for Qwen 3 32B.} Accuracy (blue, \% of DV) and token reduction (orange, \% vs.\ DV) versus confidence threshold $\tau$ on MedQA, MathQA, MedMCQA, and MMLU. $\tau=1.0$ is the DV baseline (100\% accuracy, 0\% reduction).
    }
    \label{fig:qwen3-32_cali}
\end{figure*}

The calibrated confidence threshold for each LLM on the validation set is shown in Figures~\ref{fig:llama_cali}--\ref{fig:qwen3-32_cali}. The key observation is that the accuracy--efficiency trade-off is model dependent, but most calibration curves contain a usable region where token reduction increases while accuracy remains close to the dynamic-voting reference. This supports lightweight validation-set threshold calibration: for each dataset, we select the threshold that maximizes token reduction while keeping validation accuracy within 0.5 percentage points of the best validation accuracy.

\section{Accuracy--Efficiency Trade-off Results for Additional LLMs}
\label{appendix:b}

\begin{figure*}[t]
    \centering
    \includegraphics[width=.85\linewidth]{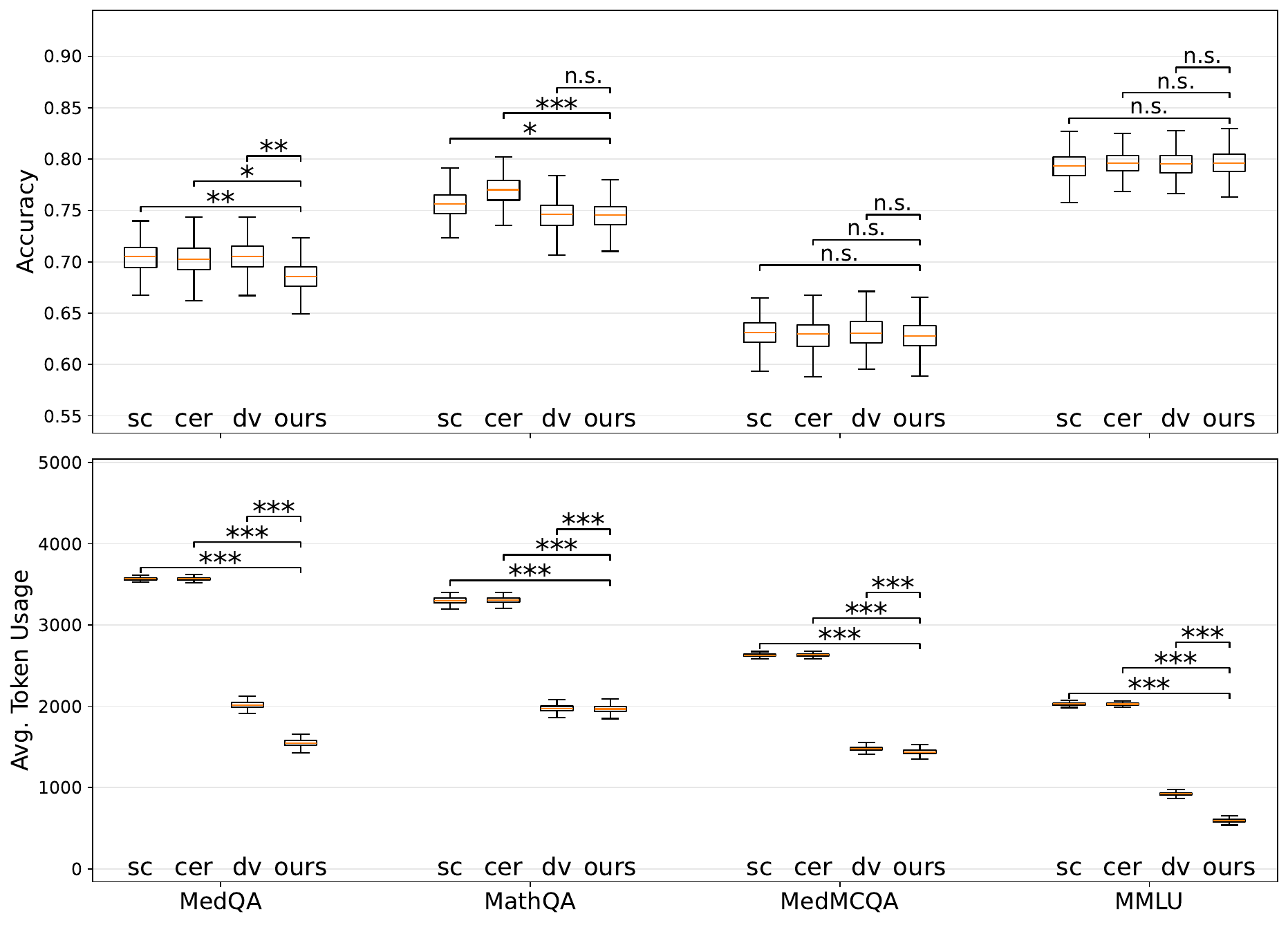}
    \caption{\textbf{Distribution of accuracy and token usage across datasets.}
Boxplots show the performance of SC, CER, DV, and Ours on MedQA, MathQA, MedMCQA, and MMLU using Llama 3.1 8B Instruct. The top panel reports accuracy, while the bottom panel reports average token usage. Statistical significance is evaluated using paired bootstrap with 2{,}000 resamples; 
$n.s.$ indicates not significant and $^{*}$ indicates $p<0.05$; $^{**}$ indicates $p<0.01$; $^{***}$ indicates $p<0.001$. Higher accuracy is better, whereas lower token usage indicates greater efficiency.}
    \label{fig:llama_boxplot}
\end{figure*}

\begin{figure*}[t]
    \centering
    \includegraphics[width=.85\linewidth]{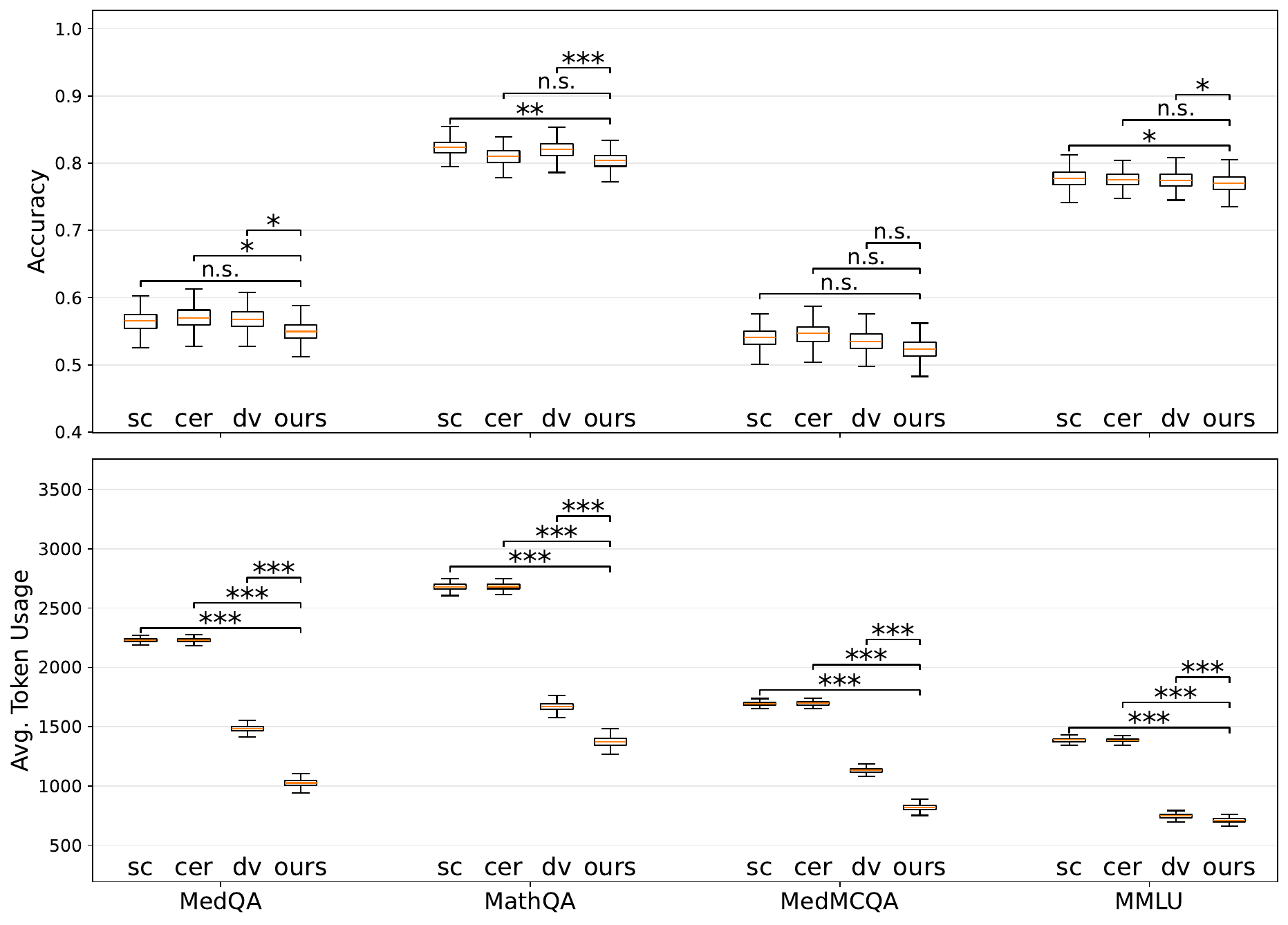}
    \caption{\textbf{Distribution of accuracy and token usage across datasets.}
Boxplots show the performance of SC, CER, DV, and Ours on MedQA, MathQA, MedMCQA, and MMLU using Qwen 2.5 7B. The top panel reports accuracy, while the bottom panel reports average token usage. Statistical significance is evaluated using paired bootstrap with 2{,}000 resamples; 
$n.s.$ indicates not significant and $^{*}$ indicates $p<0.05$; $^{**}$ indicates $p<0.01$; $^{***}$ indicates $p<0.001$. Higher accuracy is better, whereas lower token usage indicates greater efficiency.}
    \label{fig:qwen2_boxplot}
\end{figure*}

\begin{figure*}[t]
    \centering
    \includegraphics[width=.85\linewidth]{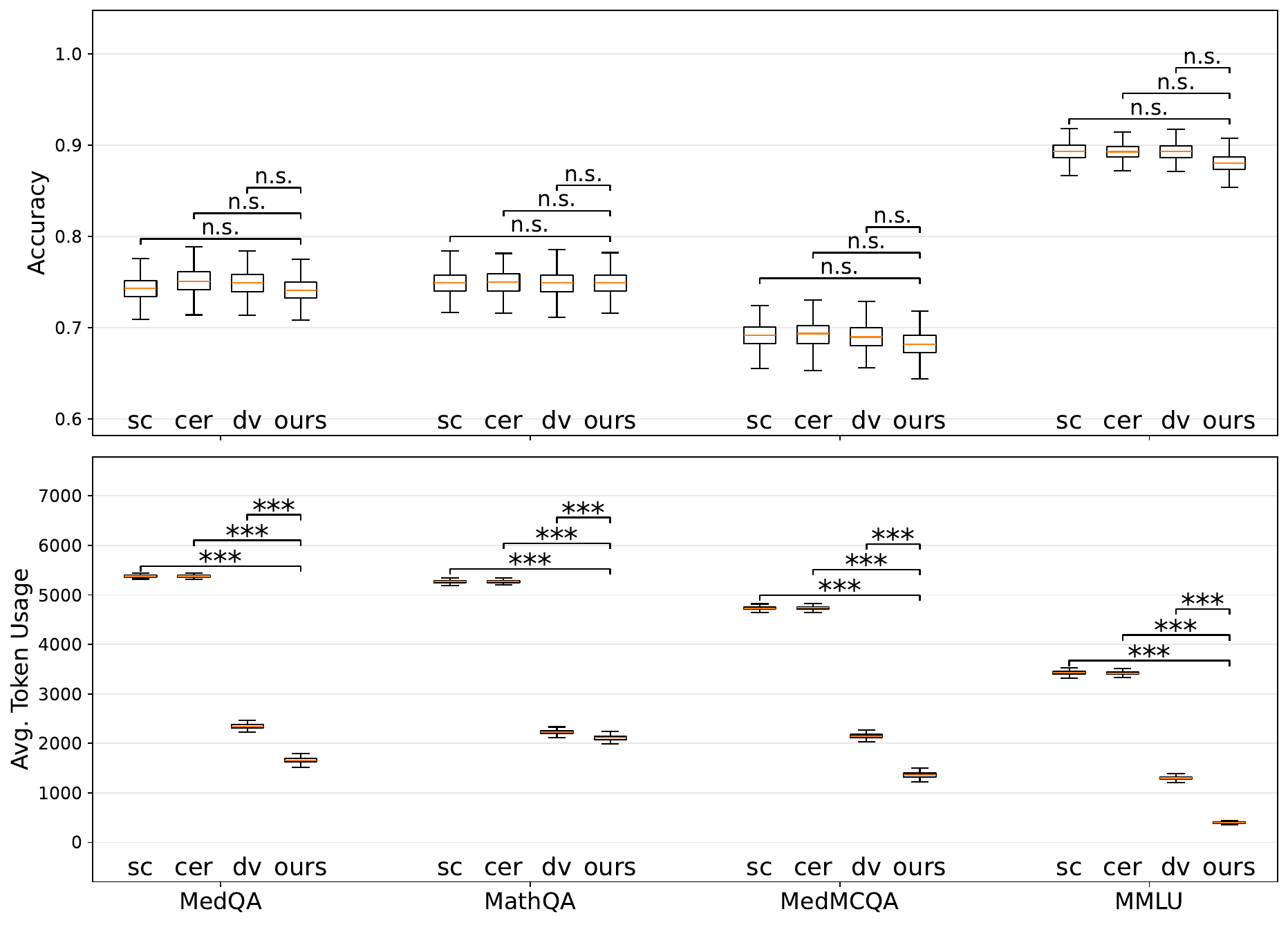}
    \caption{\textbf{Distribution of accuracy and token usage across datasets.}
Boxplots show the performance of SC, CER, DV, and Ours on MedQA, MathQA, MedMCQA, and MMLU using Qwen 3 14B. The top panel reports accuracy, while the bottom panel reports average token usage. Statistical significance is evaluated using paired bootstrap with 2{,}000 resamples; 
$n.s.$ indicates not significant and $^{*}$ indicates $p<0.05$; $^{**}$ indicates $p<0.01$; $^{***}$ indicates $p<0.001$. Higher accuracy is better, whereas lower token usage indicates greater efficiency.}
    \label{fig:qwen3-14_boxplot}
\end{figure*}

\begin{figure*}[t]
    \centering
    \includegraphics[width=.85\linewidth]{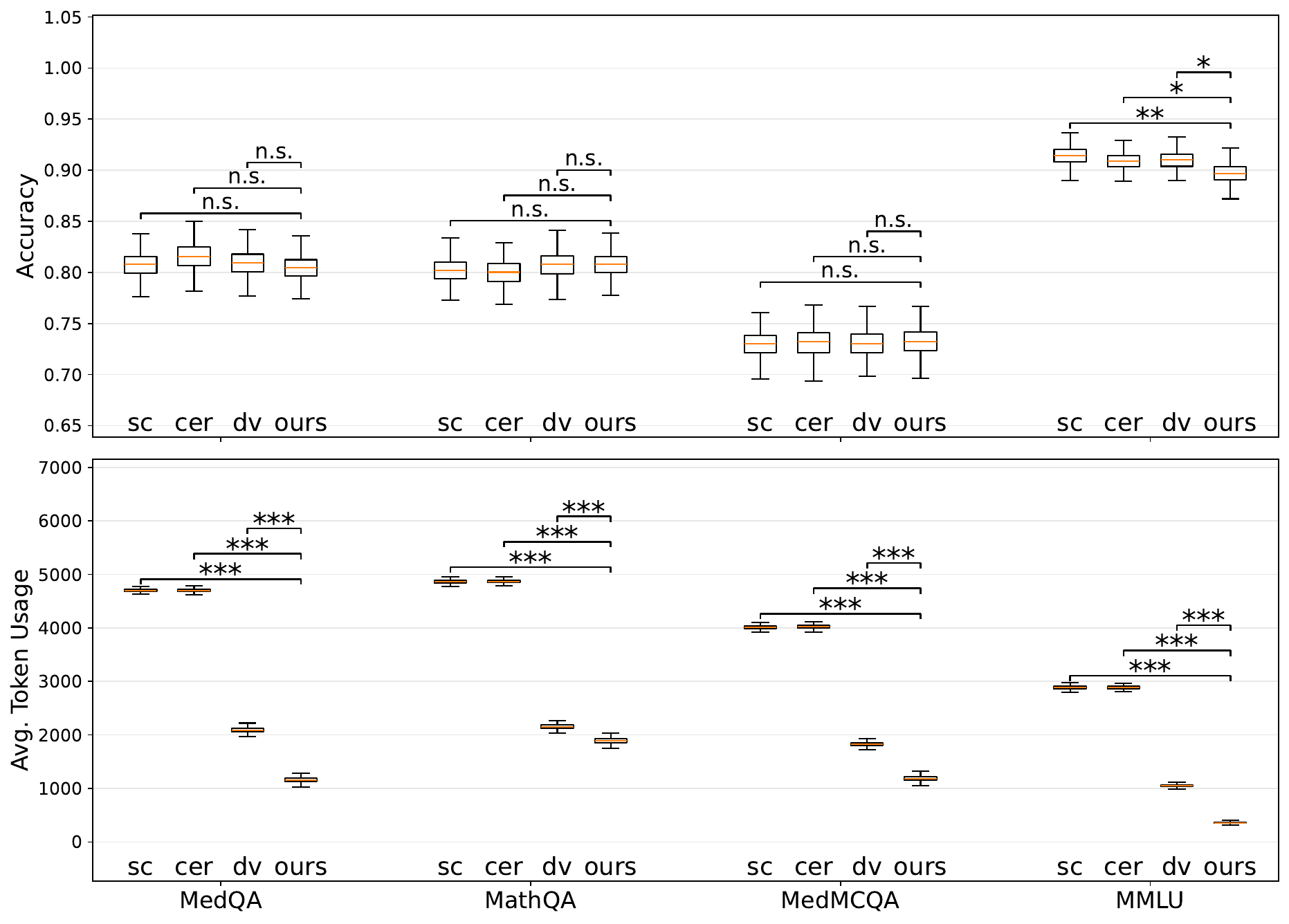}
    \caption{\textbf{Distribution of accuracy and token usage across datasets.}
Boxplots show the performance of SC, CER, DV, and Ours on MedQA, MathQA, MedMCQA, and MMLU using Qwen 3 32B. The top panel reports accuracy, while the bottom panel reports average token usage. Statistical significance is evaluated using paired bootstrap with 2{,}000 resamples; 
$n.s.$ indicates not significant and $^{*}$ indicates $p<0.05$; $^{**}$ indicates $p<0.01$; $^{***}$ indicates $p<0.001$. Higher accuracy is better, whereas lower token usage indicates greater efficiency.}
    \label{fig:qwen3-32_boxplot}
\end{figure*}

For each LLM, the decision model is trained on MedQA traces and evaluated on MedQA, MathQA, MedMCQA, and MMLU. MedQA serves as the in-domain setting, while the remaining datasets assess calibration-only transfer. Figures~\ref{fig:llama_boxplot}--\ref{fig:qwen3-32_boxplot} show that the main trend is consistent across backbones: selective sampling substantially reduces generated token usage while keeping accuracy close to multi-path reasoning baselines. The amount of token reduction varies by model and dataset, reflecting differences in the separability of confidence and entropy trajectories; noisier trajectories lead to more conservative routing, while clearer trajectories allow more single-path outputs to be accepted.

\end{document}